\documentclass[letterpaper, 10 pt, conference]{ieeeconf}  

\IEEEoverridecommandlockouts                              

\usepackage{color}
\newcommand{\ie}{i.e.\ }
\newcommand{\eg}{e.g.\ }
\newcommand{\Reffig}[1]{Figure~\ref{#1}}

\newcommand{\Refalg}[1]{Algorithm~\ref{#1}}
\newcommand{\Reftab}[1]{Table~\ref{#1}}

\usepackage{graphics} 
\usepackage{epsfig} 
\usepackage{algorithm}
\usepackage{algpseudocode}
\usepackage{amsmath}
\usepackage{subfigure}
\usepackage{booktabs}
\usepackage{multirow}
\usepackage{multicol}
\usepackage{makecell}

\title{\LARGE \bf
Real-time Multi-target Path Prediction and Planning for Autonomous Driving aided by FCN 
}

\author{Hongtu Zhou$^{1, 2, 3}$ Xinneng Yang$^{1, 2}$, Enwei Zhang$^{1, 2}$, Junqiao Zhao$^{*, 1, 2, 3}$, Lewen Cai$^{4}$, Chen Ye$^{1, 2}$, Yan Wu$^{1, 2}$
\thanks{This work is supported by the National Key Research and Development Program of China (No. 2018YFB0105103, No. 2017YFA0603104, No. 2018YFB0505400), the National Natural Science Foundation of China (No. U1764261, No. 41801335, No. 41871370), the Shanghai Science and Technology Development Foundation (No. 17DZ1100202, No. 16DZ1100701) and the Fundamental Research Funds for the Central Universities (No. 22120180095).}
\thanks{$^{1}$The Key Laboratory of Embedded System and Service Computing, Ministry of Education, Tongji University, Shanghai
        {\tt\small zhaojunqiao@tongji.edu.cn}}%
\thanks{$^{2}$Department of Computer Science and Technology, School of Electronics and Information Engineering, Tongji University, Shanghai}
\thanks{$^{3}$Institute of Intelligent Vehicle, Tongji University, Shanghai}%
\thanks{$^{4}$Department of Computer Science, Nanjing University, Nanjing}%
}

\begin{document}

\maketitle
\thispagestyle{empty}
\pagestyle{empty}

\begin{abstract}

Real-time multi-target path planning is a key issue in the field of autonomous driving.
Although multiple paths can be generated in real-time with polynomial curves, the generated paths are not flexible enough to deal with complex road scenes such as S-shaped road and unstructured scenes such as parking lots. 
Search and sampling-based methods, such as A* and RRT and their derived methods, are flexible in generating paths for these complex road environments. 
However, the existing algorithms require significant time to plan to multiple targets, which greatly limits their application in autonomous driving. 
In this paper, a real-time path planning method for multi-targets is proposed.
We train a fully convolutional neural network (FCN) to predict a path region for the target at first.
By taking the predicted path region as soft constraints, the A* algorithm is then applied to search the exact path to the target.
Experiments show that FCN can make multiple predictions in a very short time (50 times in 40ms), and the predicted path region effectively restrict the searching space for the following A* search.
Therefore, the A* can search much faster so that the multi-target path planning can be achieved in real-time (3 targets in less than 100ms).
\end{abstract}

\section{INTRODUCTION}
The path planning methods commonly used in autonomous driving include path generation based on polynomial curves and path planning methods based on sampling and search \cite{gonzalez2015review}.
Among them, the search-based and sampling-based methods, such as A*\cite{hart1968formal} and RRT \cite{lavalle1998rapidly}, can search for paths in a variety of complex road environments.
However, these methods usually require significant time to plan a path to one target.
Therefore, when paths to multiple targets are required, the time cost will be overwhelming to meet the real-time demand for autonomous driving.

In recent years, deep learning has been shown to enable semantic understanding of images, such as recognition \cite{he2016deep,huang2017densely} and segmentation \cite{zhao2017pyramid,fu2019dual}.
At the same time, deep learning has been directly used for path prediction \cite{caltagirone2017lidar}.
However, due to the unexplainable nature of the neural network, it is risky to use the results directly for autonomous driving.
Nevertheless, these studies inspired us to combine accurate search-based path planning algorithms with deep learning-based methods. 
We employed neural networks to learn and predict the path search region for multiple targets, and then adopt this as constraints to restrict the search space of A*, thus enabling real-time multi-target path planning.

First, we use an existing A*-based path planning method (TiEV A*) that has been deployed in an autonomous driving vehicle \cite{zhao2018TIEV} to create training samples fully automatically.
The path planning result of TiEV A* is dilated and represented by a binary image as the labeled ground truth.
The obstacle map generated by laser scanning, the rough referencing path, and the target point is used as input information for the network.
We employ the fully convolutional network (FCN) to learn the labeled path region and the results show a 90\% of mIOU was obtained in 2000 test samples.
On this basis, an A*-based multi-target path planning algorithm (multi-target TiEV A*) is put forward, which gives priority to search in the path region predicted by the neural network.
Because the neural network can predict the path region to target considering the obstacles and the reference path, the multi-target TiEV A* can complete path planning in about 55 ms with three different targets.
The time cost is reduced by 45\% compared to the original TiEV A* without the path prediction.
Moreover, because the predicted path region is treated as a soft constraint, the correctness of the planned path can always be guaranteed by using the A* algorithm.

\section{RELATED WORKS}
In this section, the flexible path planning methods based on searching and sampling are briefly reviewed.
Then the pioneering deep learning-based methods for path planning are also introduced. 
\subsection{Search-based and Sampling-based path planning}
The A* search algorithm proposed by \cite{hart1968formal} is one of the most popular path searching methods of high performance. 
However, the naive A* algorithm cannot be applied to autonomous driving because of the lack of angle dimension.
Hybrid A* extended the A* algorithm by searching in continuous coordinate space, which allows the modeling of vehicle motion models \cite{dolgov2008practical}.
This method was successful in path planning tasks for driving in semi-structured road environments \cite{dolgov2010path}.

Because A* is suitable for static scenes, and its performance can be degraded in cases where the scene became dynamic, the D* search algorithm was proposed \cite{stentz1994d}.
\cite{kolski2006autonomous} adopted the Field D* algorithm in global path planning for autonomous driving.
\cite{ferguson2008motion} adopted the anytime D* for path planning in unstructured road environments. 
The search-based path planning algorithms can find a path efficiently in lower-dimensional search space, \ie{2D or 3D}.
However, path planning for autonomous driving usually requires considering more dimensions, \eg{temporal information}.
As a result, the sampling-based path planning methods were proposed for planning in high dimensional space.
\cite{lavalle1998rapidly} proposed the rapid exploring random tree (RRT) algorithm to tackle such a problem.
\cite{kuwata2008motion} adopted the RRT algorithm for autonomous driving and improved its efficiency.
The algorithm was further extended to combine the closed-loop control by \cite{kuwata2009real}.
Recently, the vehicle model prediction was also integrated into the algorithm \cite{arslan2017sampling}.
Sampling-based path planning methods can find a path in high dimensional space efficiently.
However, they can easily get stuck when narrow portals present in the scene.
And both the search-based and the sampling-based methods require significant time for planning to one target.
Therefore, the real-time path planning for multi-targets cannot be achieved by using these methods.


\subsection{Deep learning-based path planning }
In recent years, deep learning is intensively explored in autonomous driving research.
However, many of the methods implemented an end-to-end paradigm by feeding in the perception information and outputting the control instructions directly, \eg{the steering angle and velocity} \cite{pfeiffer2017perception}.
However, the end-to-end methods are still unstable for autonomous driving in real road environments.

%

In \cite{caltagirone2017lidar}, the FCN is adopted to predict the path by learning from the history paths.
The laser scanning map, the historic path and the driving intentions, \ie{straight, left, right}, were used to train the FCN to predict the future paths for three possible directions, \ie{straight, left turn, right turn}.
The results proved that the FCN is capable of interpreting the road environments in the laser scanning map and can predict the feasible paths for driving. 
However, the path generated from FCN can be unstable and inaccurate.
So that it is risky to directly adopt such a path in autonomous driving.
A combined solution of deep learning and exact path planning methods is demanded.

%
\section{Path prediction with FCN}

The FCN is widely used in the semantic segmentation task of images.
In autonomous driving systems, it is often used to achieve the extraction of drivable areas and lane lines \cite{hou2019learning,tsutsui2017distantly}.
In this paper, we propose to use FCN to learn the path planning ability of TiEV A*.

\begin{figure}[htb]
\centering
\subfigure[TiEV A* planning]{\label{train.example}\includegraphics[height=2.7in]{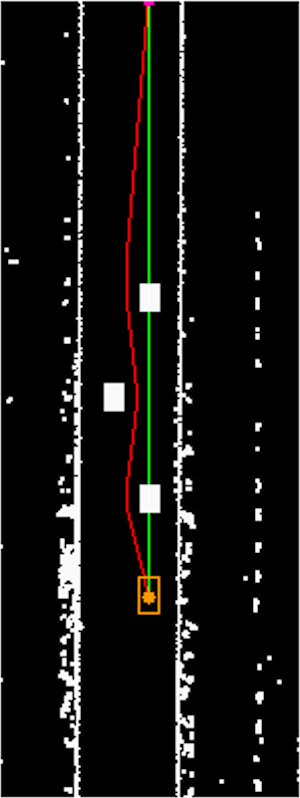}}
\subfigure[FCN input]{\label{train.input}\includegraphics[height=2.7in]{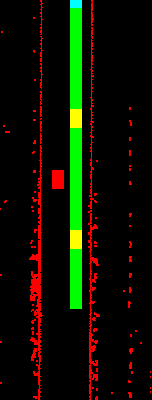}}
\subfigure[FCN label]{\label{train.label}\includegraphics[height=2.7in]{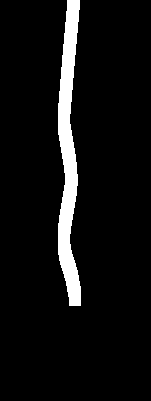}}
\caption{\subref{train.example}: The path planning in TiEV A*, where the obstacles are shown in white, the planned path is shown in red, the ego position of the vehicle is shown in orange, the target is shown by a pink dot and the reference global path is shown in green; \subref{train.input}: The generated input of the training sample from the TiEV A* planning results; \subref{train.label}: The generated label of the training sample from the TiEV A* planning results.}
\label{train}
\end{figure}

\begin{figure*}[ht]
\centering
\includegraphics[width=\textwidth]{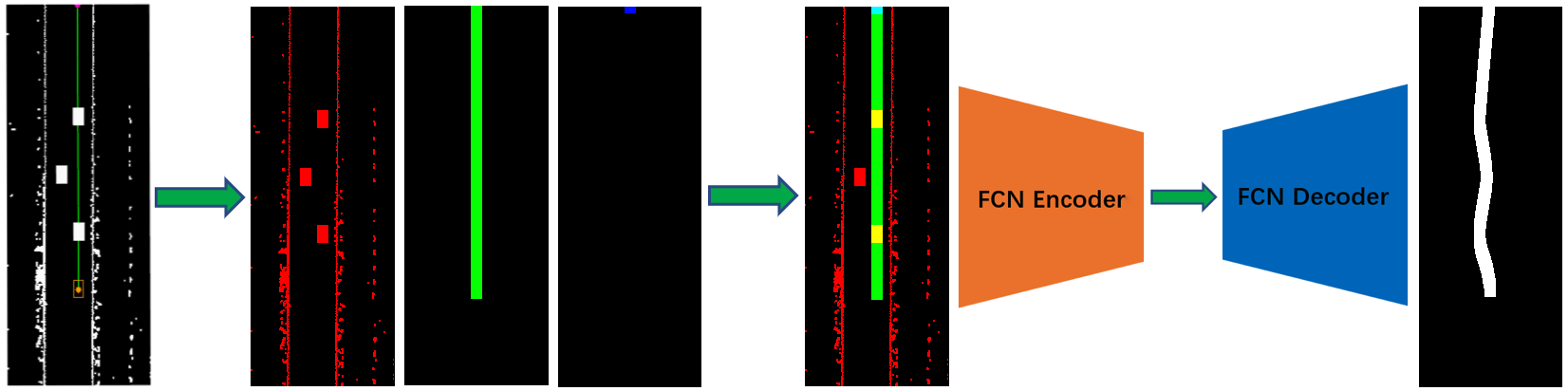}
\caption{The proposed FCN workflow: Giving the combined perception input (the first stage); The three main components are extracted (the second stage), where the obstacle map is shown in red, the reference path map is shown in green and the target map is shown in blue; These three components are merged into a three-channel image which is the input of the FCN encoder (the third stage); Finally, the FCN is trained against the path labels (the fourth stage).}
\label{TiEVfcn}
\end{figure*}

We start with the existing TiEV A* path planning method to automatically generate massive training samples. 
The perception information fed to TiEV A* includes the static obstacle map (white), the dynamic obstacle object (white), the reference path derived from the global planning (green) as shown in \Reffig{train.example}.
The TiEV A* then search the path from the ego position (orange) to the target (red dot) (\Reffig{train.example}).
We then generate the training sample out of the above planning results.
The input of the training sample is composed of three components: the obstruction region (red), the reference region (green) and the target region (blue) as shown in \Reffig{train.input}.
The point and line features are all dilated to regions to facilitate feature encoding in FCN. 
The label of the training sample is the dilated path region as shown by the binary image (\Reffig{train.lable}).
The workflow is illustrated in \Reffig{TiEVfcn}.

\subsection{Automatic training sample generation with TiEV A*}

\begin{figure}[htb]
\centering
\includegraphics[width=0.3\textwidth]{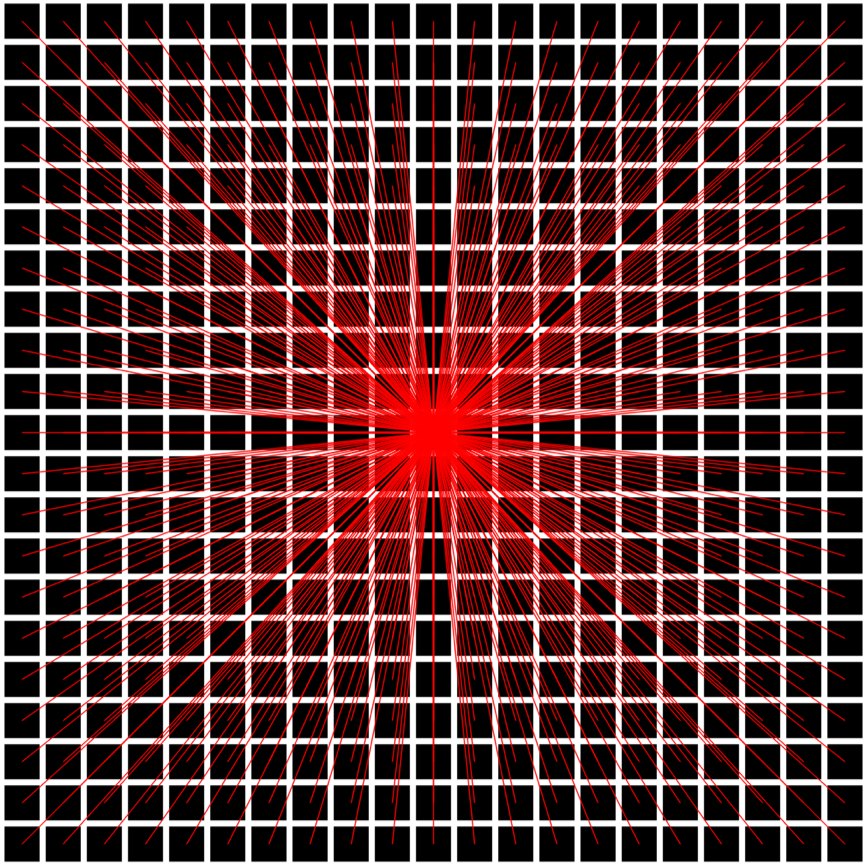}
\caption{TiEV A*'s lookup table: This lookup table is composed of 21 by 21 grids. 
Assuming the center gird is the current expanding point, this lookup table approximate 368 different directions, so it also has 368 different actions}
\label{lookuptable}
\end{figure}

TiEV A* is a grid map-based search method proposed for path planning in both structured and unstructured road environments. 
A 21 by 21 \emph{LookUpTable} was devised to approximate the multiple search angles (\Reffig{lookuptable}).
In each search step, the extended nodes are selected from the \emph{LookUpTable} within the allowed angle range, which is defined according to the vehicle velocity.
The $Cost$ of each search step is calculated by the sum of distance $\Delta Dis$ between the extended node and the current node and the angle difference $\Delta Ang$:\[ Cost = \Delta Dis + \Delta Ang .\]
The heuristics of TiEV A* is calculated by the Euclidean distance between the current node and the target.
The algorithm can search for a smooth path to a given target in a variety of complex environments, as shown by \Reffig{train.example}.

\begin{figure}[htb]
\centering
\includegraphics[width=0.5\textwidth]{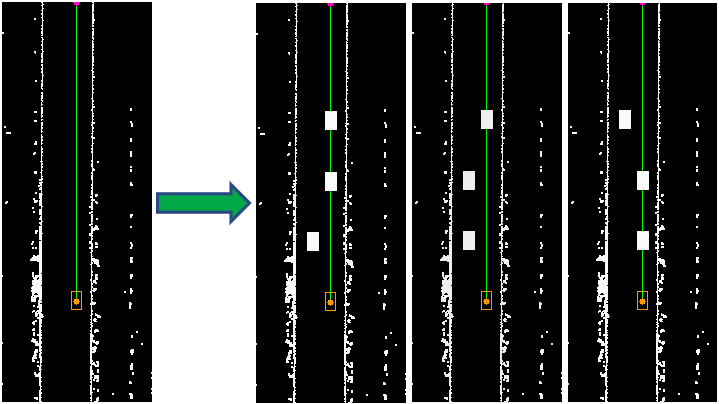}
\caption{Augmentation of the obstruction map by randomly adding simulated vehicle obstacles}
\label{obs_generate}
\end{figure}

\begin{figure}[htb]
\centering
\includegraphics[width=0.5\textwidth]{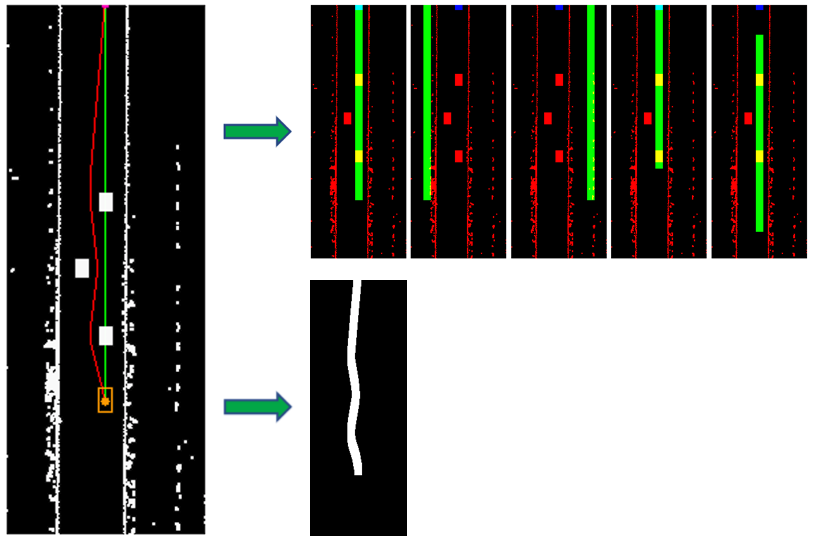}
\caption{Augmentation of the reference global path: The figure on the left represents the planning result of TiEV A*; The graph at the top right is the results of the randomly shifted reference global path; The bottom right figure is the FCN label extracted from the original planning result}
\label{generate_data}
\end{figure}

We extracted the obstacle map from hours long log data of the IVFC competition \cite{zhao2018tiev}. 
We take all the perception input of the TiEV A* and argument the data for generating a massive set of training samples.
The obstruction map is augmented by randomly adding simulated vehicle obstacles along the reference path and its parallel path to make the environment more complicated, as shown in \Reffig{obs_generate}.
We also randomly shifted the reference global path to simulated the consequences of the unstable online localization, as shown in \Reffig{generate_data}. 

%

Finally, we sampled the vertical and horizontal equidistant distances along the direction of the reference path to obtain multiple planning targets.
For each target, 5 training samples are generated automatically.

\subsection{Path learning using FCN}
In this paper, we adopt a real-time FCN implementation, ENet \cite{paszke2016enet}, as the segmentation network and made some modifications. 
ENet has the advantages of good generalization, small network model and low computational complexity. 
In our implementation, we use avg pooling to replace max-pooling in the downsampling bottleneck, and use bilinear upsampling to replace max unpooling in the upsampling bottleneck, which is able to hold more information. 
We use Adam Optimizer with a batch size of 100, an initial learning rate of 0.0005 and a weight decay of 0.0002. 
For the learning rate schedule, we use the learning rate warmup strategy suggested by \cite{he2019bag} and Cosine Learning Rate Decay.

Assume the total number of batches is $T$, then at batch $i$, the learning rate $l_i$ is computed as:
 \[l_i = \frac{1}{2}*(1+\cos{\frac{i*\pi}{T}})*l_{base}\]
We divide the generated 13,485 training samples into 11485 images for training and 2000 images for testing. 
The model is trained for 300 epochs on the train set and is evaluated based on the test set.

\section{Real-time Multi-target path planning}
On the basis that the FCN is able to predict the path region to the target in a very short time, we propose a real-time multi-target path planning method, which is composed of the FCN path region prediction and A*-based path searching.

%

\subsection{Improving TiEV A* with FCN's predicted path region}
We introduce the predicted path region by the FCN as soft constrains for the following A*-based search.
If an expanding node of A* is located in the predicted path region, we multiply $Cost$ and heuristics $h$ by a coefficient $w$ between 0 to 1 (0.15 in our experiment), which gives a high priority to searching in predicted path region.
The A* search will then reach the target in a faster fashion compared to the plain A*.
The improved TiEV A* algorithm pseudo-code is shown in \Refalg{imprTiEVa*}.

\begin{algorithm}[htb]
\caption{Improved TiEV A* Algorithm}
\begin{algorithmic}[1]
\Require
$LidarMap; Speed; LookUpTable; Road_{fcn}; T_{list}$
\Ensure
$PlanedPath$
\State initialize $S, TimeLimit, Q, time, PlanedPath, w$
\State $AngLimit \gets getAngLimit(Speed)$
\State $T \gets getTarget(Road_{fcn}, T_{list})$
\State $Q.push(S)$
\Repeat
\State $P_{now} \gets Q.pop()$
\If  {$P_{now} = T$ \textbf{or} $P_{now} = null$} \textbf{break}
\EndIf
\State $P_{children} \gets getChildren(AngLimit)$
\For{each $P_{next} \in P_{children}$}
\State $Cost \gets \Delta Dis + \Delta Ang$
\If{($P_{now}.g + Cost < P_{next}.g$)}
\State $P_{next}.parent \gets P_{now}$
\State $P_{next}.g \gets P_{now}.g + Cost$
\State $P_{next}.h \gets EuclideanDistance(P_{next}, T)$
\If{$P_{next} \in Road_{fcn}$}
\State $P_{next}.g \gets P_{now}.g + Cost*w$
\State $P_{next}.h \gets P_{next}.h*w$
\EndIf
\State $P_{next}.f \gets P_{next}.g + P_{next}.h$
\State $Q.push(P_{next})$
\EndIf
\EndFor
\State $time \gets getTime()$
\Until{($time \geq TimeLimit$)}
\State $PlanedPath \gets backTrack(T_{target})$ \\
\Return $PlanedPath$
\end{algorithmic}
\label{imprTiEVa*}
\end{algorithm}

\subsection{Real-time multi-target path planning}
Based on the efficient FCN and TiEV A* combination, the multi-target path planning can then be implemented.
Given the obstacles map and the reference path, multiple targets are obtained by sampling at horizontal and vertical equidistant distances along the direction of the reference path.
The FCN first predicts the path region for every target in a batch.
The path planning for each target is then performed by the improved TiEV A*.
Since FCN can predict a batch of path regions in a short time (50 images under 40ms in our experiment), and the improved TiEV A* requires much lesser time for planning a path to each target, the path planning for multiple targets can then be achieved in real-time.
The pseudo-code of the proposed algorithm as shown in \Refalg{multitarget}.

\begin{algorithm}[htb]
\caption{Multi-target Path Planning Aided by FCN}
\begin{algorithmic}[1]
\Require
$LidarMap; RefPath; FCN$
\Ensure
$PathList$
\State initialize $PathList$, $PicList$
\State $RefPathMap \gets getPathMap(RefPath)$
\State $T_{list} \gets getTargets(RefPath, LidarMap)$
\For{each $T \in T_{list}$}
\State $TMap \gets getTMap(T)$
\State $Pic \gets combine(LidarMap, RefPathMap, TMap)$
\State $PicList.push(Pic)$
\EndFor
\State $Roads_{fcn} \gets FCN.predict(PicList)$
\For{each $Road_{fcn} \in Roads_{fcn}$}
\State get $PlanedPath$ using Improved TiEV A*
\State $PathList.push(PlanedPath)$
\EndFor \\
\Return $PathList$
\end{algorithmic}
\label{multitarget}
\end{algorithm}


\section{Experiments}
\subsection{Path prediction result}
\begin{figure}[ht]
\centering
\subfigure[FCN Inputs]{\label{fcninput}
\includegraphics[width=0.09\textwidth]{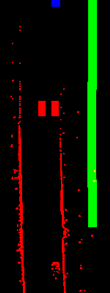}\hspace{0.001\textwidth}
\includegraphics[width=0.09\textwidth]{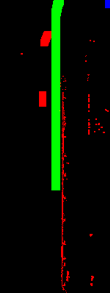}\hspace{0.001\textwidth}
\includegraphics[width=0.09\textwidth]{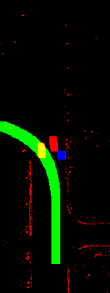}\hspace{0.001\textwidth}
\includegraphics[width=0.09\textwidth]{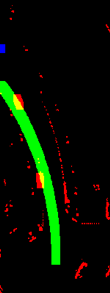}\hspace{0.001\textwidth}
\includegraphics[width=0.09\textwidth]{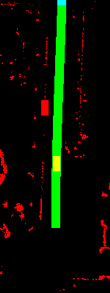}}

\subfigure[FCN's Prediction]{\label{fcnpredict}
\includegraphics[width=0.09\textwidth]{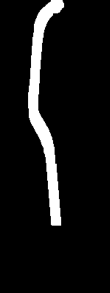}\hspace{0.001\textwidth}
\includegraphics[width=0.09\textwidth]{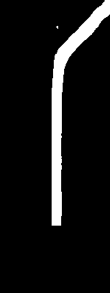}\hspace{0.001\textwidth}
\includegraphics[width=0.09\textwidth]{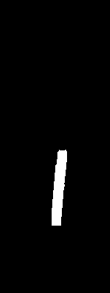}\hspace{0.001\textwidth}
\includegraphics[width=0.09\textwidth]{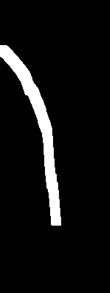}\hspace{0.001\textwidth}
\includegraphics[width=0.09\textwidth]{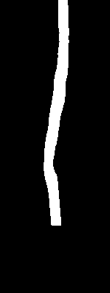}}

\subfigure[Groundthuth]{\label{fcngt}
\includegraphics[width=0.09\textwidth]{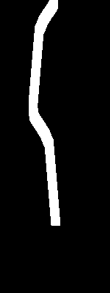}\hspace{0.001\textwidth}
\includegraphics[width=0.09\textwidth]{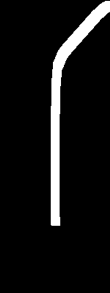}\hspace{0.001\textwidth}
\includegraphics[width=0.09\textwidth]{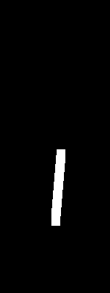}\hspace{0.001\textwidth}
\includegraphics[width=0.09\textwidth]{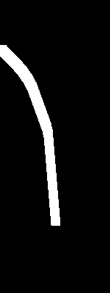}\hspace{0.001\textwidth}
\includegraphics[width=0.09\textwidth]{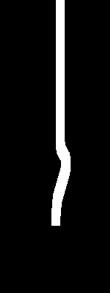}}

\caption{Qualitative examples of the segmentation produced by FCN. \subref{fcninput} the input image. \subref{fcnpredict} network outputs. \subref{fcngt} Ground Truth.}
\label{fcnresult}
\end{figure}

In this section, we evaluate the path prediction performance of the neural network.
Five representative examples of the path prediction are shown in \Reffig{fcnresult} for qualitative evaluation (one for each column).

In all the examples except the fifth one, the reference global path (green) was shifted and only the rough direction for the path planning can be inferred from the input \Reffig{fcnresult} a).
However, giving the target, it shows that FCN succeeded in predicting obstacle-free paths for all the cases (\Reffig{fcnresult} b)).
This indicts that the neural network cannot be disturbed by the inaccurate localization, represented by the shifted reference global paths.
In the fifth example, the neural network interestingly chose a different obstacle avoiding path from the ground truth.
The predicted path region surrounds the vehicle from the left side rather than from the right.
This can be interpreted that the neural network is capable of mimic the real path searching process.
The quantitative evaluation from the 2000 testing samples shows the FCN-based path prediction achieved 90.19\% mIoU (mean Intersection over Union).

\subsection{Path planning result}
We then tested the performance of the improved TiEV A*.
Firstly, we compared the time cost and the number of expanded nodes between the original TiEV A* and the improved TiEV A* in the case where only one target is planned (as shown in \Reffig{train.example}).

\begin{figure}[ht]
\centering
\subfigure[TiEV A*]{\label{TiEVastar}\includegraphics[height=2.7in]{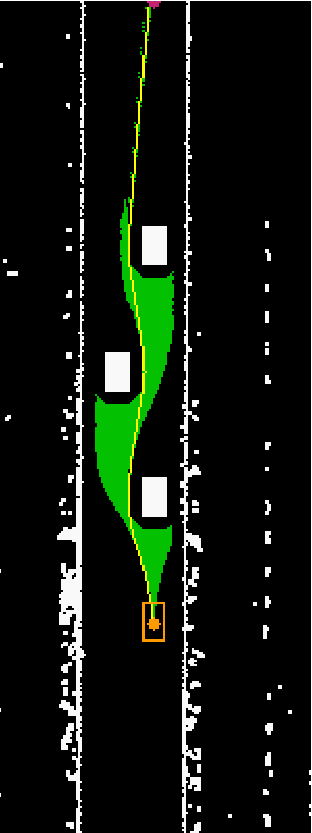}}
\subfigure[Improved TiEV A*]{\label{impTiEVastar}\includegraphics[height=2.7in]{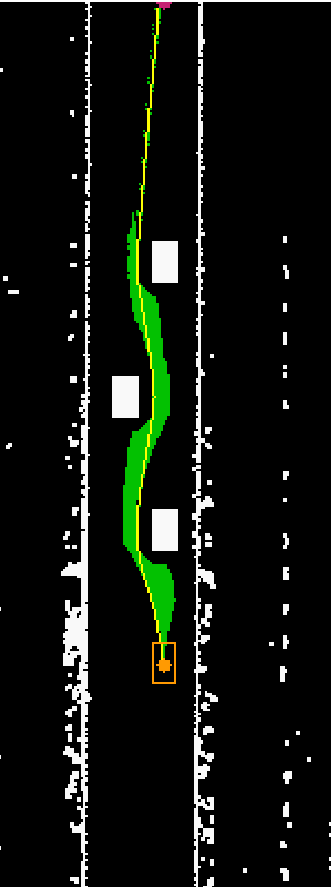}}
\caption{The comparison of the search space of the original TiEV A* \subref{TiEVastar} and the FCN aided improved TiEV A* \subref{impTiEVastar}, where the search space is shown in green.}
\label{onetargetfig}
\end{figure}

The performance is evaluated on an industrial PC equipped with a quad-core Intel i7 3610 CPU at 3.3GHz and an Nvidia GTX1060 GPU.
\Reftab{onetargettable} shows the experimental results of both search time and the number of expanded nodes, \Reffig{onetargetfig} shows a comparison of the expansion space of the two.
It should be noted that the actual search space of A* is three-dimensional $(x, y, \theta)$, but \Reffig{onetargetfig} shows only the two-dimensional space $(x, y)$.
As can be seen from \Reftab{onetargettable}, the improved TiEV A* is almost half the number of TiEV A*, both in search time and in terms of the number of extended nodes.
As can be seen from \Reffig{onetargetfig}, due to the restriction of FCN predicted path region, the expanded nodes (green region) of improved TiEV A* are much less than those of the original TiEV A*.

\begin{table}[ht]
\setlength{\abovecaptionskip}{-5pt}
\caption{The comparison of the performances of one-target path planning.}
\label{onetargettable}
\begin{center}
\begin{tabular}{ccc}
\toprule
\specialrule{0em}{3pt}{3pt}
Method & Planning Time(ms) & Searching Steps \\
\specialrule{0em}{3pt}{3pt}
\midrule
\specialrule{0em}{3pt}{3pt}
TiEV A* & 29.128 & 73224 \\
\specialrule{0em}{3pt}{3pt}
Improved TiEV A* & 15.431 & 34675 \\
\specialrule{0em}{3pt}{3pt}
\bottomrule
\end{tabular}
\end{center}
\end{table}

\begin{figure}[ht]
\centering
\subfigure[Multi-target path planning with TiEV A*]{\label{multi_TiEV}\includegraphics[width=0.5\textwidth]{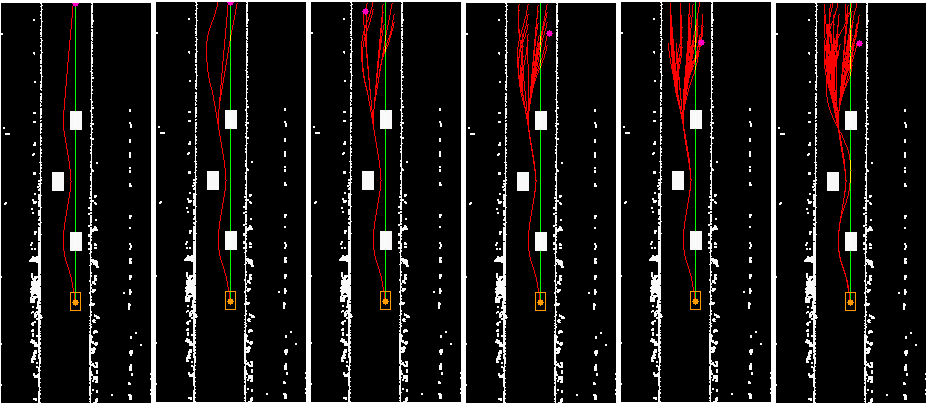}}
\subfigure[Multi-target path planning with improved TiEV A* aided by FCN]{\label{multi_fcn}\includegraphics[width=0.5\textwidth]{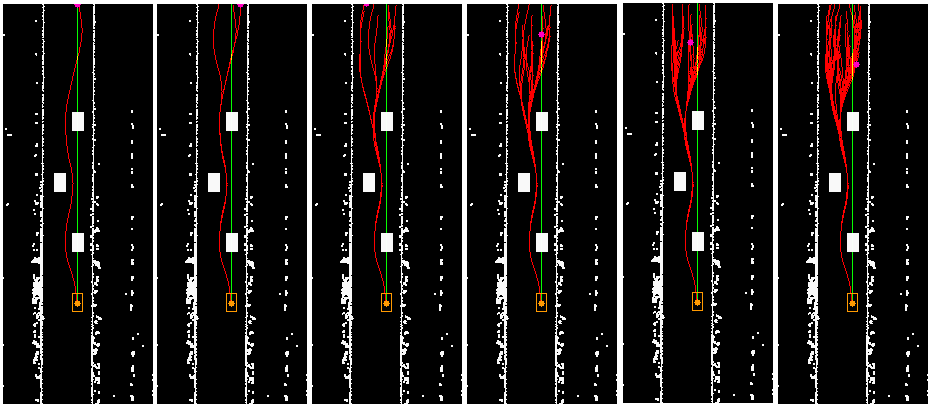}}
\caption{The comparison of the path planning results between the original TiEV A* and the improved TiEV A*, where the number of targets is 1, 3, 10, 20, 30, 50 from left to right.}
\label{multitargetfig}
\end{figure}

We then compared the time cost and the number of successfully searched targets between the two for multiple targets.
All the statistics including the time cost for path prediction and planning are shown in \Reftab{multitargettable}. 
The resulting paths of the two methods in different target numbers are also shown in \Reffig{multitargetfig}.
As can be seen from \Reftab{multitargettable}: 
\begin{enumerate}
\item The time cost of FCN prediction increases slowly with the increasing number of targets.
\item The time cost of path planning of the improved TiEV A* is approximately half of that of the TiEV A*.
\item The success rate of the improved TiEV A* plan is also a bit higher than that of the TiEV A*.
\end{enumerate}
As can be seen in \Reffig{multitargetfig}, the quality of paths of the improved TiEV A* is similar to that of the original TiEV A*.

\begin{table}[ht]
\setlength{\abovecaptionskip}{-5pt}
\caption{The comparison of the performances of multi-target path planning}
\label{multitargettable}
\begin{center}
\begin{tabular}{cccccc}
\toprule
\specialrule{0em}{3pt}{3pt}
\multirow{2}{*}{\makecell[c]{Target \\ Numbers}} & \multicolumn{2}{c}{TiEV A*} & \multicolumn{2}{c}{Improved TiEV A*} & \multirow{2}{*}{\makecell[c]{FCN \\ Time(ms)}} \\ \cmidrule(lr){2-3} \cmidrule(lr){4-5} & Time(ms) & Success & Time(ms) & Success \\
\specialrule{0em}{3pt}{3pt}
\midrule
\specialrule{0em}{3pt}{3pt}
1 & 29.128 & 1 & 15.431 & 1 & 16.32 \\
\specialrule{0em}{3pt}{3pt}
3 & 98.871 & 3 & 54.446 & 3 & 17.85 \\
\specialrule{0em}{3pt}{3pt}
7 & 212.849 & 7 & 118.947 & 7 & 19.16 \\
\specialrule{0em}{3pt}{3pt}
10 & 295.527 & 10 & 176.991 & 10 & 20.42 \\
\specialrule{0em}{3pt}{3pt}
15 & 521.167 & 15 & 288.653 & 15 & 24.08 \\
\specialrule{0em}{3pt}{3pt}
20 & 595.910 & 20 & 363.247 & 20 & 26.92 \\
\specialrule{0em}{3pt}{3pt}
30 & 1027.837 & 30 & 594.900 & 30 & 32.69 \\
\specialrule{0em}{3pt}{3pt}
40 & 1602.486 & 38 & 978.453 & 40 & 35.00 \\
\specialrule{0em}{3pt}{3pt}
50 & 2026.185 & 49 & 992.514 & 50 & 39.28 \\
\specialrule{0em}{3pt}{3pt}
\bottomrule
\end{tabular}
\end{center}
\end{table}

\section{Conclusion}
This paper proposes to use FCN to predict the path region to a target and integrate the predicted path region as a soft constraint for the A* path planning algorithm.
The experiments show the proposed FCN-aided method can significantly improve the efficiency of the original A* path search algorithm so that the multi-target path planning can be achieved in real-time.
We also discovered that FCN can not only learn the ability of path planning of the A* algorithm but is also immune from the disturbance of inaccurate or even erroneous localization results.  
A limitation of this study is that the present TiEV A* can only plan to one target at a time, which offsets the benefits brought by the batch path prediction of FCN.
Therefore, a parallel version of TiEV A* should be proposed.
In future work, it might be interesting to see how the FCN-aided method can be transparent to other path planning methods.
And it is worth exploring how to train the neural network to help select the best path in a multi-target path planning task.





\bibliographystyle{IEEEtrans}
\bibliography{IEEEabrv,ref}

\end{document}